%% file: main_arxiv.tex
\documentclass[letterpaper]{article} % DO NOT CHANGE THIS
\usepackage[draft]{aaai2026}  % DO NOT CHANGE THIS
\usepackage{times}  % DO NOT CHANGE THIS
\usepackage{helvet}  % DO NOT CHANGE THIS
\usepackage{courier}  % DO NOT CHANGE THIS
\usepackage[hyphens]{url}  % DO NOT CHANGE THIS
\usepackage{graphicx} % DO NOT CHANGE THIS
\usepackage{natbib}  % DO NOT CHANGE THIS AND DO NOT ADD ANY OPTIONS TO IT
\usepackage{caption} % DO NOT CHANGE THIS AND DO NOT ADD ANY OPTIONS TO IT
\usepackage{listings}
\DeclareCaptionStyle{ruled}{labelfont=normalfont,labelsep=colon,strut=off} % DO NOT CHANGE THIS
\makeatletter
\@ifundefined{isChecklistMainFile}{
  \newif\ifreproStandalone
  \reproStandalonetrue
}{
  \newif\ifreproStandalone
  \reproStandalonefalse
}
\makeatother

\usepackage{bm}
\usepackage{capt-of}
\usepackage{mathtools}
\usepackage{siunitx}
\usepackage{amsfonts}
\usepackage{scrextend}
\usepackage{xspace}
\usepackage{xcolor}
\usepackage{multirow}
\usepackage{amsthm}
\usepackage{thmtools}
\usepackage{algorithm,algpseudocode}
\usepackage[export]{adjustbox}
\usepackage{url}
\usepackage{lineno}
\usepackage{pifont}
\usepackage{booktabs}
\usepackage{soul}
\usepackage{tikz}
\usetikzlibrary{arrows.meta}
\usetikzlibrary{decorations.pathmorphing}
\usepackage{subfig}

\newcounter{algsubstate}

\newtheorem*{theorem*}{Theorem}

\renewcommand{\eqref}[1]{\mbox{Eq.~(\ref{#1})}}

\makeatletter
\DeclareRobustCommand\onedot{\futurelet\@let@token\@onedot}
\def\@onedot{\ifx\@let@token.\else.\null\fi\xspace}

\def\ie{\emph{i.e}\onedot}

\def\etal{\emph{et al}\onedot}
\makeatother

\newcolumntype{L}[1]{>{\raggedright\arraybackslash}p{#1}}
\newcolumntype{C}[1]{>{\centering\arraybackslash}p{#1}}
\newcolumntype{R}[1]{>{\raggedleft\arraybackslash}p{#1}}

\title{CurveFlow: Curvature-Guided Flow Matching for Image Generation}
\author{
    Yan Luo\textsuperscript{\rm 1}, Drake Du\textsuperscript{\rm 1,2}, Hao Huang\textsuperscript{\rm 3}, Yi Fang\textsuperscript{\rm 3}, Mengyu Wang\textsuperscript{\rm 1,2} 
}
\affiliations{
    \textsuperscript{\rm 1}Harvard AI and Robotics Lab, Harvard University\\
    \textsuperscript{\rm 2}Kempner Institute for the Study of Natural and Artificial Intelligence, Harvard University\\
    \textsuperscript{\rm 3}Embodied AI and Robotics (AIR) Lab, New York University Abu Dhabi\\
    yluo16@meei.harvard.edu, drakedu@college.harvard.edu, \{hh1811, yfang\}@nyu.edu, mengyu\_wang@meei.harvard.edu
}

\usepackage{bibentry}
\begin{document}

\maketitle


% \REVISION{1. revise intro; 2. revise experiments; 3. move some sections to suppl; 4. add a section of computational time; 5. LoRA}

\input{sec/0_abstract}    
\input{sec/1_intro}

\input{sec/2_related}

\input{sec/3_method}
\input{sec/4_experiment}
\input{sec/5_conclusion}
\bibliography{main}


% \input{sec/9_checklist}

\end{document}

%% file: sec/0_abstract.tex
\begin{abstract}
Existing rectified flow models are based on linear trajectories between data and noise distributions. This linearity enforces zero curvature, which can inadvertently force the image generation process through low-probability regions of the data manifold. A key question remains underexplored: how does the curvature of these trajectories correlate with the semantic alignment between generated images and their corresponding captions, i.e., instructional compliance? To address this, we introduce CurveFlow, a novel flow matching framework designed to learn smooth, non-linear trajectories by directly incorporating curvature guidance into the flow path. Our method features a robust curvature regularization technique that penalizes abrupt changes in the trajectory's intrinsic dynamics.Extensive experiments on MS COCO 2014 and 2017 demonstrate that CurveFlow achieves state-of-the-art performance in text-to-image generation, significantly outperforming both standard rectified flow variants and other non-linear baselines like Rectified Diffusion. The improvements are especially evident in semantic consistency metrics such as BLEU, METEOR, ROUGE, and CLAIR. This confirms that our curvature-aware modeling substantially enhances the model's ability to faithfully follow complex instructions while simultaneously maintaining high image quality. The code is made publicly available at
\url{https://github.com/anonymous4science/CurveFlow}. 
\end{abstract}

%% file: sec/1_intro.tex
\section{Introduction}
\label{sec:intro}

\begin{figure}[!t]
\centering
\includegraphics[width=.9\linewidth]{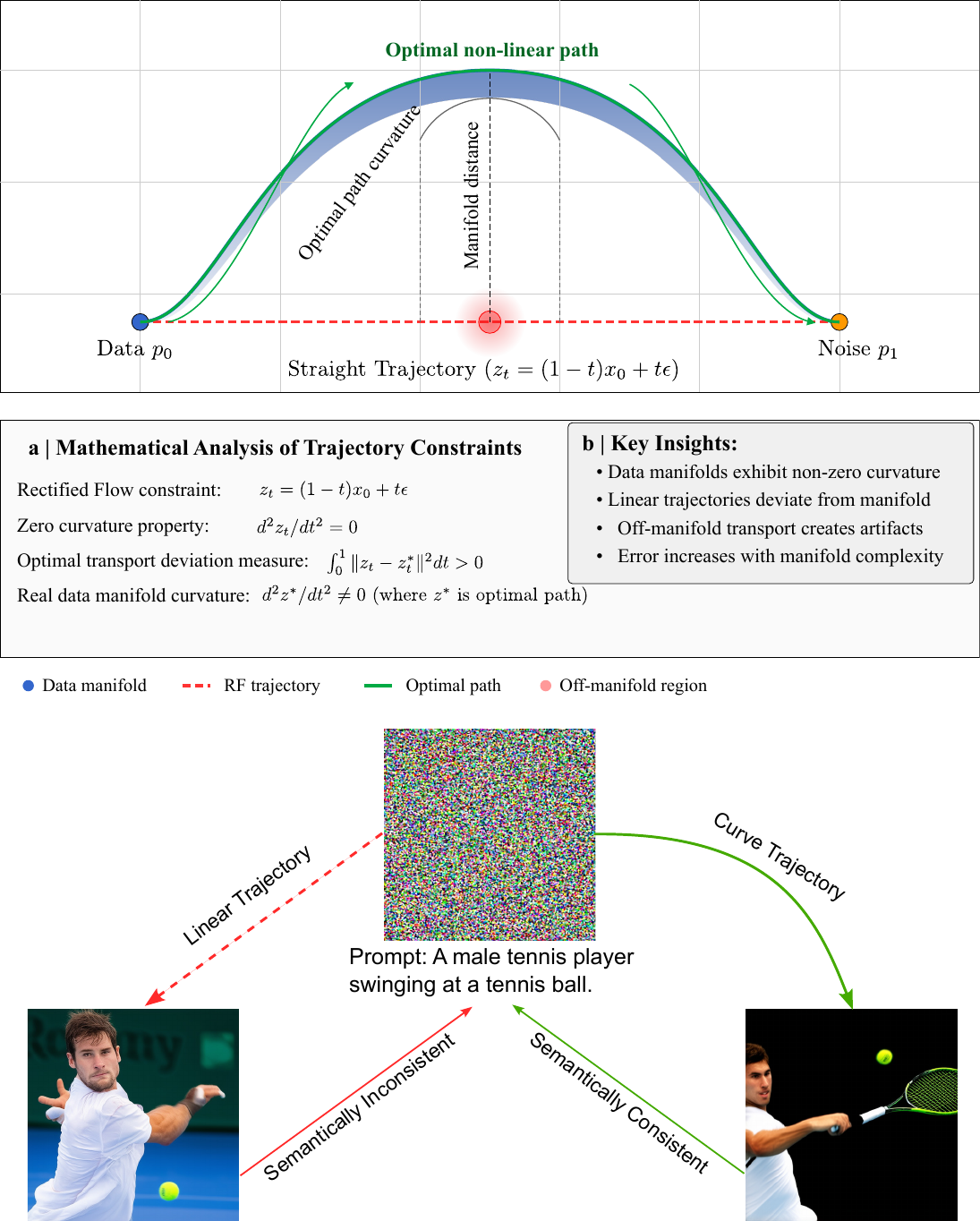}
\caption{Limitations of linear trajectory hypothesis in rectified flows (RFs). Rectified flow's linear trajectory hypothesis could break down when modeling complex data distributions. Real data typically resides on non-linear manifolds (blue curve), while rectified flows enforce linear paths (red dashed line) that deviate from the optimal transport path (green), leading to generation artifacts. As a result, the images generated by rectified flows may not accurately capture the semantics of the prompts.}
\label{fig:teasing}
\end{figure}

Generative modeling has undergone a transformative shift with the emergence of diffusion-based approaches, which excel at generating high-quality images by progressively transforming noise into structured data \cite{ho2020denoising}. However, diffusion models, such as Denoising Diffusion Probabilistic Models (DDPMs)~\cite{ho2020denoising}, rely on a highly temporally non-linear noise addition forward process and require many steps to sample high-quality data due to its stochastic inference process. Recent advances, such as Denoising Diffusion Implicit Models (DDIMs)~\cite{song2020denoising} and DMP-Solver~\cite{lu2022dpm}, have speeded up inference by modeling the inference process as a deterministic ordinary differential equation (ODE). Flow Matching (FM) models~\cite{lipman2023flow,albergo2023building} learn a continuous flow of ODE that maps samples from a simple distribution (typically a standard Gaussian) to the data distribution with learned velocity fields. Compared to diffusion models such as DDPM and DDIM, flow matching models define the forward process through explicitly hypothesized trajectories, rather than a highly temporally non-linear noise addition forward process. The most popular flow matching model, Rectified Flow (RF)~\cite{albergo2023building}, explicitly uses linear interpolation between data samples and noise (Gaussian samples) as the initial forward process based on straight trajectory hypothesis, and then learns a continuous ODE-based flow to refine and improve these trajectories. Due to the simplified linear trajectory hypothesis, rectified flow models can achieve faster training and sampling times compared to traditional diffusion models while maintaining comparable or even superior image generation quality~\cite{lipman2023flow, liu2022flow, esser2024scaling}.
%but this choice of path is flexible. 
%More specifically, RF constructs a straight-line trajectory between data and noise distributions, eliminating the need for simulating a forward stochastic differential equation process, offering fast sampling—potentially in just a few steps or even a single step—by integrating the learned velocity field~\cite{esser2024scaling}. This approach has demonstrated impressive results in image generation, offering a computationally efficient alternative to the multi-step sampling procedures used in many diffusion models \cite{lipman2023flow, liu2022flow}.

However, we observe that the linear trajectory hypothesis underlying rectified flow introduces a significant limitation. The rectified flow assumes a linear path with zero curvature, which may not suffice to capture the intricate transformations between the image and noise distributions \cite{esser2024scaling}, as shown in Figure \ref{fig:teasing}. Real-world image data, such as those in ImageNet \cite{deng2009imagenet} or MS COCO \cite{lin2014microsoft}, often lie in non-linear manifolds characterized by diverse textures, styles, and semantic structures \cite{esser2024scaling}. The rigid linearity of rectified flow trajectories, where the second derivative \( \frac{d^2 z_t}{dt^2} = 0 \), forces the model to traverse potentially low-probability regions in the joint image and noise distribution space, risking artifacts or reduced sample fidelity \cite{ho2020denoising}. Although rectified flow offers greater training and inference efficiency compared to diffusion models, it is less flexible in capturing the underlying geometry of complex data, a limitation contrasted by the expressive capability of diffusion models~\cite{rombach2022high, ho2020denoising}. The recent findings in \cite{rectified_diffusion_2025} confirm our hypothesis that this inflexibility arises directly from the strictly linear trajectory. Embracing these crucial non-linear transformations is not just vital for high-fidelity image generation, but it may also unlock a deeper, more nuanced understanding of visual data, significantly enhancing the semantic alignment of visual content with text modalities, i.e., instructional compliance.
% confirm our hypothesis that such inflexibility arises from the assumption of a strictly linear trajectory, which oversimplifies the inherently non-linear transformations between image and noise distributions essential for high-fidelity image generation. Moreover, better visual fidelity may unleash the ability of the flow models to understand more nuanced understanding of image data, allowing to the semantic alignment of visual content with text modalities.

To address this limitation, we propose \textit{CurveFlow}, a novel curvature-guided flow matching model that introduces flexibility to handle non-linear image-to-noise transitions and vice versa. By incorporating curvature information into the flow trajectory, parameterized as \( z_t = a_\phi(t) x_0 + b_\psi(t) \epsilon \) with learned coefficients \( a_\phi(t) \) and \( b_\psi(t) \), our approach guides the flow matching model to learn a smoother transformation trajectory between noise and image distributions.
Specifically, we propose CurveFlow, a curvature-guided flow framework that connects data to noise via smooth, geometrically-informed trajectories by leveraging the curvature of parametric paths defined by scaling functions $a_\phi(t)$ and $b_\psi(t)$. To ensure stable and robust training, we introduce a novel curvature regularization that penalizes rapid changes in the coefficient functions' dynamics, effectively minimizing the intrinsic "turning" behavior of the flow without relying on noisy empirical averages. Extensive experimental results on MS COCO 2014 and 2017 \cite{lin2014microsoft} verify the effectiveness of our proposed method. The contributions of this work can be summarized as:
%bridges the gap between the rigid linear paths of RF and the more flexible but computationally intensive sampling procedures of diffusion models \cite{liu2022flow, ho2020denoising}.
% Specifically, we propose a new curvature-guided regularization term (see Section~\ref{sec:training_objective}). In addition, to adaptively allocate computational resources across different timesteps during training, we introduce a novel dynamic noise scheduling to focus learning on data regions with high trajectory curvature (see Section~\ref{subsec:dynamic_noise}). We aim to advance the state-of-the-art in image generation, offering a trajectory-flexible and inference-efficient framework to generate high-fidelity images. Comprehensive experimental results on MS COCO 2014 and 2017 \cite{lin2014microsoft} verify the effectiveness of our proposed method. The contributions of this work can be summarized as:
\begin{itemize}
    \item We propose \textit{CurveFlow} guide by curvature information to learn a smoother transformation trajectory of mapping between image and noise.
    % \textit{CurveFlow} achieves this through a curvature regularization term and curvature-based dynamic noise scheduling.
    \item We conduct comprehensive experiments on MS COCO 2014 and 2017 for text-to-image generation. The evaluation results demonstrate that \textit{CurveFlow} achieves state-of-the-art performance in text-to-image generation.
    % \item In particular, \textit{CurveFlow} substantially improves instructional compliance in text-to-image generation quantified by text similarity scores including BLEU, METEOR, ROUGE, and CLAIR.
    \item \textit{CurveFlow} significantly enhances instructional compliance in text-to-image generation, as evidenced by improved text similarity scores across metrics such as BLEU, METEOR, ROUGE, and CLAIR.
\end{itemize}

%% file: sec/2_related.tex
\section{Related Work}
\label{sec:related}

Generative modeling has evolved significantly with the introduction of diffusion models, which have become a cornerstone for high-quality image synthesis. Denoising Diffusion Probabilistic Models (DDPMs)~\cite{ho2020denoising} pioneer this paradigm by iteratively adding Gaussian noise into data samples through a Markov chain and reversing this process for image generation. While effective, DDPMs suffer from slow sampling due to multi-step denoising. Subsequent advancements, such as Denoising Diffusion Implicit Models (DDIMs)~\cite{song2020denoising}, introduce deterministic sampling to accelerate inference while retaining quality. Latent Diffusion Models (LDMs)~\cite{rombach2022high} further improve efficiency by operating in a latent space, enabling scalable training on complex datasets like LAION-5B~\cite{schuhmann2022laion}. Other notable efforts include perception prioritized training of diffusion models~\cite{choi2023perception}, which enhances perceptual quality, and consistency models~\cite{song2023consistency}, which reduce sampling steps by enforcing trajectory consistency. Despite these innovations, diffusion models generally require numerous denoising steps that increase computational overhead and error accumulation~\cite{esser2024scaling}.

Flow matching generative models offer a compelling alternative by constructing continuous transformations between noise and data distributions with forward process modeled by explicitly hypothesized trajectory, which results in more efficient training and inference. Rectified Flow, a popular flow matching model, introduced by Liu \etal~\cite{liu2022flow}, formulates this forward process as a straight-line trajectory~\cite{lipman2023flow}. Recent works have built upon rectified flow to enhance its capabilities. For instance, Lee \etal~\cite{improving_rf_2024} optimize training stability, while Rout \etal~\cite{semantic_rf_2025} improve one-round training of rectified flows. Lukoianov \etal~\cite{score_distillation_2024} integrate rectified flow with distillation techniques for faster image generation. Other advancements have been made, including Yin \etal~\cite{one_step_diffusion_2024} achieves single-step inference with distillation and evaluates on LAION Aesthetics \cite{schuhmann2022laion}, InstaFlow~\cite{liu2023instaflow} turns Stable Diffusion~\cite{rombach2022high} into a one-step model with ultra-fast inference, Piecewise Rectified Flow~\cite{yan2024perflow} accelerates diffusion models by dividing the sampling process into different time windows and straightens the trajectories in these windows. Most recently, Wang \etal~\cite{rectified_diffusion_2025} challenge the necessity of strict linearity of flow trajectories.
%
%Yin et al., \cite{improved_distillation_2024} also refines this approach, while 
%Latent Consistency Models (LCMs)~\cite{luo2023latent} and LCM-LoRA~\cite{luo2023lcm_lora} further accelerate Stable Diffusion-based generation,
%
% The primary advantage of flow matching models, particularly rectified flow, lies in their computational efficiency. By enforcing a straight trajectory from noise \( p_1 = \mathcal{N}(0, I) \) to data \( p_0 \), rectified flow significantly reduces the number of integration steps required for inference—sometimes to a single step—compared with DDPM and DDIM~\cite{esser2024scaling, liu2022flow}. This efficiency stems from the linear interpolation \( z_t = (1 - t) x_0 + t \epsilon \), which simplifies the transformation trajectory of the underlying ODE. %Moreover, the Conditional Flow Matching (CFM) objective~\cite{lipman2023flow} enables tractable training without simulating intermediate distributions, a significant improvement over the simulation-heavy training of traditional diffusion models.
%
% Despite its advantages, the linear trajectory hypothesis in rectified flow imposes limitations in capturing complex, nonlinear data transformations. 
As real-world datasets, such as high-resolution images, often lie on intricate manifolds with non-zero curvature~\cite{esser2024scaling}, a linear path from data samples to noises may traverse low-probability regions, leading to suboptimal sample quality or artifacts~\cite{rectified_diffusion_2025}. 
%
% This rigidity contrasts with diffusion models like DDPMs and DDIMs, which adapt to curved paths that better align with data geometry, albeit at higher computational cost. 
%
% Prior works like \cite{rectified_diffusion_2025} also argue that strict linearity may oversimplify the generative process, particularly for semantically rich data. Nevertheless, it does not explore the impact of curvature on semantic alignment between the generated images and the original captions.
% The linearity constraint, while simplifying computation, sacrifices expressiveness in high-fidelity and diverse image generation, which motivates our proposed curvature-guide flow matching model \textit{CurveFlow}.
Although the prior work \cite{rectified_diffusion_2025} argues that strict linearity may oversimplify the generative process, particularly for semantically rich data, it does not investigate how curvature influences the semantic alignment between generated images and their corresponding captions, i.e., instructional compliance.

%% file: sec/3_method.tex
\section{Background \& Problem Statement}
\label{subsec:background_rf}
Diffusion models transform noise into image data by inverting a forward process that temporally nonlinearly adds noise to data samples \cite{ho2020denoising}. Rectified Flow (RF) introduced in~\cite{liu2022flow,albergo2023building,lipman2023flow} offers an alternative formulation that establishes a straight-line connection between the data distribution \( p_0 \) and the noise distribution \( p_1 = \mathcal{N}(0, I) \). Specifically, the forward process in RF is defined as a linear interpolation:
\begin{align}
z_t = (1 - t) x_0 + t \epsilon\enspace, \quad t \in [0, 1]\enspace,
\label{eqn:rf_trajectory}
\end{align}
where \( x_0 \sim p_0 \) represents a data sample, \( \epsilon \sim p_1 \) is Gaussian noise, and \( t \) is the time parameter. This straight path simplifies the transformation compared to the trajectories of diffusion models~\cite{ho2020denoising,rombach2022high}). The corresponding velocity field \( v_\Theta(z_t, t) \) is learned to match the time derivative of the straight trajectory, enabling a Flow Matching training objective:
\begin{align}
\mathcal{L}_{\text{FM}} = \mathbb{E}_{t, p_t(z|\epsilon), p(\epsilon)} \left\| v_\Theta(z, t) - u_t(z|\epsilon) \right\|_2^2\enspace,
\label{eqn:cfm_loss}
\end{align}
where \( u_t(z|\epsilon) = \epsilon - x_0 \) is the velocity field derived from the forward process, and \( \Theta \) denotes the neural network parameters. This linear path, a hallmark of RF, enables fast sampling—potentially in a single step—due to its straightforward integration properties~\cite{esser2024scaling}.
However, the linearity of this trajectory imposes a significant geometric constraint: \textit{it enforces zero curvature}. To be specific, we consider the second derivative of the trajectory with respect to time, $\frac{d^2 z_t}{dt^2} = \frac{d}{dt} \left( \frac{d z_t}{dt} \right) = \frac{d}{dt} (\epsilon - x_0) = 0$,
%
% \begin{align}
% \frac{d^2 z_t}{dt^2} = \frac{d}{dt} \left( \frac{d z_t}{dt} \right) = \frac{d}{dt} (\epsilon - x_0) = 0\enspace,
% \label{eqn:zero_curvature}
% \end{align}
%
since \( \epsilon - x_0 \) is constant. This vanishing second derivative confirms that RF trajectories are perfectly straight, a property that simplifies computation but introduces a critical limitation when modeling real-world complex data. 
% To quantify this limitation, one can consider the scenario where the true data-to-noise mapping involves non-linear transitions. The linear assumption in RF introduces a discretization bias, as it cannot adapt to curvature inherent in the transition trajectory between noise distribution $p_1$ and data distribution \( p_0 \). We can express this error as the cumulative deviation of the trajectory’s curvature from what a curve path would require. For a general path, the curvature-related error can be approximated by:
% %
% \begin{align}
% \mathcal{E} \propto \int_0^1 \left\| \frac{d^2 z_t}{dt^2} \right\| dt = \int_0^1 \left\| \frac{\partial v_\Theta}{\partial z_t} \cdot v_\Theta \right\| dt\enspace,
% \label{eqn:curvature_error}
% \end{align}
% %
% where \( \frac{d^2 z_t}{dt^2} = \frac{\partial v_\Theta}{\partial z_t} \cdot v_\Theta \) reflects the acceleration induced by changes in the velocity field. In RF, this integral evaluates to zero due to Equation~\ref{eqn:zero_curvature}, implying no error under the linear assumption. Paradoxically, this ``zero error'' comes at the cost of reduced expressiveness, as the model lacks the flexibility to adapt to non-linear manifold structures. 

This geometric rigidity not only risks traversing low-probability regions—potentially degrading sample fidelity—but also limits the model's ability to preserve fine-grained semantic correspondences during generation. In text-to-image tasks, where precise alignment between visual content and textual instructions is crucial, such limitations can manifest as hallucinations, missing attributes, or incorrect compositions. While prior work~\cite{esser2024scaling} demonstrates RF’s effectiveness in text-to-image synthesis through tailored noise sampling, the strict linearity may still hinder full capture of the semantic complexity embedded in natural image-text pairs. Thus, enabling curvature-aware trajectories is not merely a geometric refinement. It is a necessary step toward improving both sample quality and instructional compliance.

\section{Curvature-Guided Flow}

\begin{figure}[!t]
\centering
\includegraphics[width=.9\linewidth]{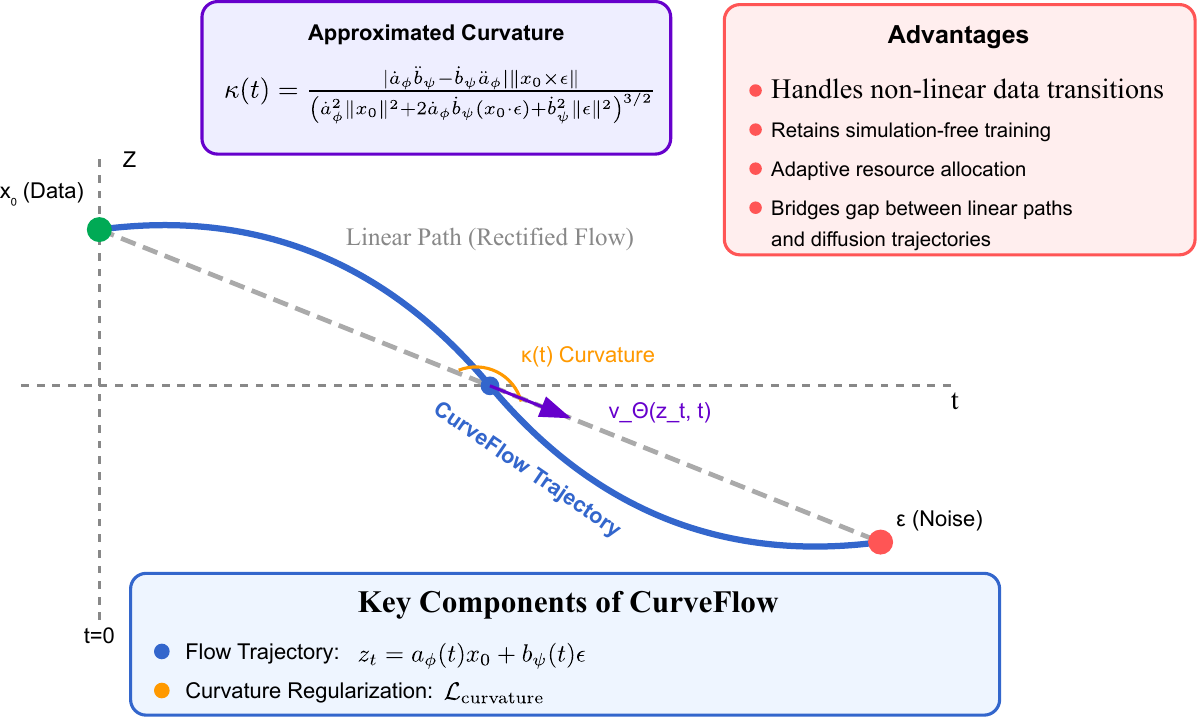}
\caption{Schematic view of \textit{CurveFlow}. The diagram illustrates how \textit{CurveFlow} establishes curve trajectories between data samples ($x_0$) and Gaussian noise ($\epsilon$), contrasting with the linear paths of Rectified Flow (dashed line). The trajectory is defined by $z_t = a_\phi(t)x_0 + b_\psi(t)\epsilon$, where curvature $\kappa(t)$ measures the deviation from linearity. Key innovations include: (1) a curvature-aware training objective $\mathcal{L}_{\text{Curve-FM}}$ that aligns the velocity field with curve paths, and (2) curvature regularization $\mathcal{L}_{\text{curvature}}$ that controls path geometry. These components together enable \textit{CurveFlow} to handle non-linear data transitions while retaining simulation-free training advantages.}
\label{fig:schematic}
\end{figure}

In this section, we propose \textit{CurveFlow}, a curvature-guided flow that connects data samples and Gaussian noises. An overview of \textit{CurveFlow} is shown in Figure \ref{fig:schematic}. We also describe new loss functions to learn \textit{CurveFlow} and a refined approach to adjust noise scheduling dynamically.

\subsection{Curvature of Trajectory}
For a parametric curve \( z(t) \) spanned by vectors \( x_0 \) and \( \epsilon \), the curvature \( \kappa(t) \) is generally defined as $\kappa(t) = \frac{\|\dot{z}(t) \times \ddot{z}(t)\|}{\|\dot{z}(t)\|^3}$,
%
% \begin{align}
% \kappa(t) = \frac{\|\dot{z}(t) \times \ddot{z}(t)\|}{\|\dot{z}(t)\|^3}\enspace,
% \label{eqn:curvature_def}
% \end{align}
%
where \( \dot{z}(t) \) is the velocity and \( \ddot{z}(t) \) is the acceleration.

\subsection{Flow Trajectory with Curvature}

We propose \textit{CurveFlow} trajectory defined as:
\begin{align}
z_t = a_\phi(t) x_0 + b_\psi(t) \epsilon\enspace,
\label{eqn:gen_trajectory}
\end{align}
with the coefficients \( a_\phi(t) \) and \( b_\psi(t) \) satisfying boundary conditions \( a_\phi(0) = 1 \), \( b_\psi(0) = 0 \), \( a_\phi(1) = 0 \), \( b_\psi(1) = 1 \). The velocity and acceleration are:
\begin{itemize}
    \item \textit{Velocity}: $\dot{z}_t = \dot{a}_\phi(t) x_0 + \dot{b}_\psi(t) \epsilon$,
    % \begin{align}
    % \dot{z}_t = \dot{a}_\phi(t) x_0 + \dot{b}_\psi(t) \epsilon\enspace,
    % \end{align}
    \item \textit{Acceleration}: $\ddot{z}_t = \ddot{a}_\phi(t) x_0 + \ddot{b}_\psi(t) \epsilon$,
    % \begin{align}
    % \ddot{z}_t = \ddot{a}_\phi(t) x_0 + \ddot{b}_\psi(t) \epsilon\enspace,
    % \end{align}
\end{itemize}
where \( \dot{a}_\phi(t) = \frac{da_\phi}{dt} \), \( \ddot{a}_\phi(t) = \frac{d^2 a_\phi}{dt^2} \), and similarly for \( b_\psi(t) \).

The cross product term becomes:
\[
\dot{z}_t \times \ddot{z}_t = (\dot{a}_\phi \ddot{b}_\psi - \dot{b}_\psi \ddot{a}_\phi) (x_0 \times \epsilon)\enspace,
\]
therefore:
\[
\|\dot{z}_t \times \ddot{z}_t\| = |\dot{a}_\phi \ddot{b}_\psi - \dot{b}_\psi \ddot{a}_\phi| \|x_0 \times \epsilon\|.
\]
The speed is:
\[
\|\dot{z}_t\|^2 = \dot{a}_\phi^2 \|x_0\|^2 + 2 \dot{a}_\phi \dot{b}_\psi (x_0 \cdot \epsilon) + \dot{b}_\psi^2 \|\epsilon\|^2.
\]
Thus, the curvature for a specific flow trajectory is:
\begin{align}
\kappa(t) = \frac{|\dot{a}_\phi \ddot{b}_\psi - \dot{b}_\psi \ddot{a}_\phi| \|x_0 \times \epsilon\|}{\left( \dot{a}_\phi^2 \|x_0\|^2 + 2 \dot{a}_\phi \dot{b}_\psi (x_0 \cdot \epsilon) + \dot{b}_\psi^2 \|\epsilon\|^2 \right)^{3/2}}.
\label{eqn:curv_def_orig}
\end{align}

As this expression depends on \( x_0 \) and \( \epsilon \), which are randomly sampled, direct regularization can introduce high variance. In the following, we introduce a more robust and stable curvature regularization based on the time-dependent coefficients.

\subsection{Training Objective with Robust Curvature Regularization}
\label{sec:training_objective_improved}
Similar to prior Flow Matching works, the data term in our training objective focuses on aligning the velocity field \( v_\Theta(z_t, t) \) with a curve path defined by \( \dot{a}_\phi(t) x_0 + \dot{b}_\psi(t) \epsilon \), as defined in \( \mathcal{L}_{\text{Curve-FM}} \):
\begin{align}
\mathcal{L}_{\text{Curve-FM}} = \mathbb{E}_{t, \epsilon} \left\| v_\Theta(z_t, t) - \left( \dot{a}_\phi(t) x_0 + \dot{b}_\psi(t) \epsilon \right) \right\|^2_2\enspace.
\label{eqn:loss_CurvA_CFM_improved}
\end{align}

To enhance learning robustness and prevent the generation of random noise, we introduce a new curvature regularization term, \( \mathcal{L}_{\text{robust\_curvature}} \), that focuses on the intrinsic properties of the scaling functions \( a_\phi(t) \) and \( b_\psi(t) \). Instead of relying on the empirical average of the full curvature expression, which can be noisy and dependent on specific samples of \( x_0 \) and \( \epsilon \), we directly penalize the "turning" behavior inherent in the coefficient functions.

Specifically, we propose to regularize the magnitude of the determinant term \( |\dot{a}_\phi \ddot{b}_\psi - \dot{b}_\psi \ddot{a}_\phi| \), which directly contributes to the numerator of the curvature and captures the rate of change in the "direction" of the flow in the $(a,b)$ parameter space. A large value for this term implies a rapid change in the relative contributions of $x_0$ and $\epsilon$, potentially leading to unstable trajectories.

The robust curvature regularization term is defined as:
\begin{align}
\mathcal{L}_{\text{robust\_curvature}} = \lambda \int_0^1 \left( \dot{a}_\phi(t) \ddot{b}_\psi(t) - \dot{b}_\psi(t) \ddot{a}_\phi(t) \right)^2 dt\enspace,
\label{eqn:loss_robust_curvature_obj}
\end{align}
where \( \lambda \) is a pre-defined hyperparameter controlling the strength of the regularization. This term directly encourages smoother and more predictable evolution of the coefficients \( a_\phi(t) \) and \( b_\psi(t) \), without dependence on specific \( x_0 \) and \( \epsilon \) samples. This promotes stability in the learned flow by penalizing excessive "twisting" in the underlying scaling functions.

The integral in Equation \ref{eqn:loss_robust_curvature_obj} is approximated as a Riemann sum over a fixed grid of \( M \) points (e.g., \( M = 1000 \)) per training step, computed separately from batch sampling. This allows gradients to propagate through \( a_\phi \) and \( b_\psi \).

To compute the derivatives \( \dot{a}_\phi \), \( \dot{b}_\psi \), \( \ddot{a}_\phi \), and \( \ddot{b}_\psi \), we employ numerical differentiation over a uniform grid. The interval \( [0, 1] \) is divided into \( M \) equal sub-intervals, with \( t_i = i / M \) for \( i = 0, 1, \dots, M \) and step size \( \Delta t = 1 / M \). First-order derivatives are approximated via central differences:
\begin{align*}
\dot{a}_\phi(t_i) &\approx \frac{a_\phi(t_{i+1}) - a_\phi(t_{i-1})}{2 \Delta t}\enspace, \\
\dot{b}_\psi(t_i) &\approx \frac{b_\psi(t_{i+1}) - b_\psi(t_{i-1})}{2 \Delta t}\enspace.
\end{align*}
Similarly, second-order derivatives are approximated as:
\begin{align*}
\ddot{a}_\phi(t_i) &\approx \frac{a_\phi(t_{i+1}) - 2 a_\phi(t_i) + a_\phi(t_{i-1})}{(\Delta t)^2}\enspace, \\
\ddot{b}_\psi(t_i) &\approx \frac{b_\psi(t_{i+1}) - 2 b_\psi(t_i) + b_\psi(t_{i-1})}{(\Delta t)^2}\enspace.
\end{align*}
These approximations enable numerical evaluation of the robust curvature regularization term across the grid, promoting stable learning by controlling the complexity of the coefficient functions.

The total loss function would then be:
\begin{align}
\mathcal{L}_{\text{total}} = \mathcal{L}_{\text{Curve-FM}} + \mathcal{L}_{\text{robust\_curvature}}\enspace.
\end{align}

%% file: sec/4_experiment.tex
\section{Experiment \& Analysis}

\noindent\textbf{Experimental Setup}.
To evaluate the proposed method, we conduct experiments by fine-tuning the rectified flow model \cite{esser2024scaling} with Low-Rank Adaptation (LoRA) \cite{hu2022lora}. The rectified flow model is obtained from the Hugging Face repository.
% To fine-tune the rectified flow model using SimpleTuner, we conduct the training process with the following parameters: The model is initialized from the stabilityai/stable-diffusion-3.5-large checkpoint, employing Low-Rank Adaptation (LoRA) for efficient parameter updates. Following SimpleTuner’s recommendation, we add standard LoRA adapters (rank 16) to MMDiT’s Transformer, introducing roughly 21 million trainable parameters. 
To fine-tune the rectified flow model using SimpleTuner, we initialize the model from the stabilityai/stable-diffusion-3.5-large checkpoint. Following SimpleTuner’s recommendations, we add standard LoRA adapters with a rank of 16 to the MMDiT Transformer, introducing approximately 21 million trainable parameters.
Training is conducted over 100 epochs with a batch size of 16, utilizing the AdamW optimizer with a learning rate of 1e-5 and a polynomial learning rate scheduler, including a warmup phase of 100 steps. Mixed precision training in bfloat16 (bf16) is enabled to optimize computational efficiency. Gradient checkpointing is activated to manage memory usage effectively. The coefficient networks $ a_\phi(t) $ and $ b_\psi(t) $ are implemented as 3-layer MLPs with a 64-dimensional intermediate feature space for simplicity. All hyperparameters can be found in our codebase repository.

\begin{table*}[t]
\centering
% \captionsetup{width=.8\linewidth}
\scriptsize
\caption{Performance of generated images produced by various loss reweighting methods on the validation set (5K samples) of COCO17 on the evaluation metrics for image-to-text semantics consistency. The CLAIR scores \cite{chan2023clair} are graded by ChatGPT 3.5 Turbo. BLIP v2 \cite{li2023blip} is used to generate captions from the images produced by each model.
}
% \vspace{-2ex}
\label{tab:perf_caption_coco17}
\adjustbox{width=1\textwidth}{
\begin{tabular}{lccccccccc}
\toprule
\textbf{Method} & \textbf{BLEU-1} {$\uparrow$} & \textbf{BLEU-2} {$\uparrow$} & \textbf{BLEU-3} {$\uparrow$} & \textbf{BLEU-4} {$\uparrow$} & \textbf{METEOR} {$\uparrow$} & \textbf{ROUGE-1} {$\uparrow$} & \textbf{ROUGE-2} {$\uparrow$} & \textbf{ROUGE-L} {$\uparrow$} & \textbf{CLAIR} {$\uparrow$}  \\
\cmidrule(lr){1-1} \cmidrule(lr){2-2} \cmidrule(lr){3-3} \cmidrule(lr){4-4} \cmidrule(lr){5-5} \cmidrule(lr){6-6} \cmidrule(lr){7-7} \cmidrule(lr){8-8} \cmidrule(lr){9-9} \cmidrule(lr){10-10}
% Rectified Diffusion & 28.36  & 18.73 & 12.18 & 7.80 & 28.97 & 37.59 & 13.89 & 33.75 & 56.79 \\
Rectified Diffusion & 29.44 & 18.78 & 12.47 & 8.15 & 29.19 & 37.90 & 14.15 & 34.18 & 61.08 \\
% 30.36
RF w/o Reweighting & 25.60 & 16.21 & 10.96 & 7.35 & 27.58 & 38.56 & 14.72 & 34.66 & 62.69 \\
RF w/ LogNorm      & 25.31 & 16.13 & 11.01 & 7.49 & 27.57 & 38.24 & 14.71 & 34.42 & 62.77 \\
RF w/ ModeSample   & 24.87 & 15.81 & 10.81 & 7.36 & 27.29 & 38.17 & 14.57 & 34.42 & 62.31 \\
RF w/ CosMap       & 25.20 & 15.98 & 10.92 & 7.45 & 27.46 & 38.38 & 14.67 & 34.57 & 62.56 \\
% CurveFlow     & \textbf{30.50} & \textbf{19.45} & \textbf{13.19} & \textbf{8.90} & \textbf{30.52} & \textbf{40.17} & \textbf{15.83} & \textbf{36.13} & \textbf{63.81} \\
CurveFlow     & \textbf{29.54} & \textbf{18.93} & \textbf{12.89} & \textbf{8.69} & \textbf{29.93} & \textbf{39.76} & \textbf{15.85} & \textbf{35.94} & \textbf{64.26} \\
\bottomrule
\end{tabular}}
\end{table*}

\begin{table*}[t]
\centering
% \captionsetup{width=.8\linewidth}
\scriptsize
\caption{Performance of generated images produced by various loss reweighting methods on the validation set (30K samples) of COCO14 on the evaluation metrics for image-to-text semantics consistency.
}
% \vspace{-2ex}
\label{tab:perf_caption_coco14}
\adjustbox{width=1\textwidth}{
\begin{tabular}{lccccccccc}
\toprule
\textbf{Method} & \textbf{BLEU-1} {$\uparrow$} & \textbf{BLEU-2} {$\uparrow$} & \textbf{BLEU-3} {$\uparrow$} & \textbf{BLEU-4} {$\uparrow$} & \textbf{METEOR} {$\uparrow$} & \textbf{ROUGE-1} {$\uparrow$} & \textbf{ROUGE-2} {$\uparrow$} & \textbf{ROUGE-L} {$\uparrow$} & \textbf{CLAIR} {$\uparrow$}  \\
\cmidrule(lr){1-1} \cmidrule(lr){2-2} \cmidrule(lr){3-3} \cmidrule(lr){4-4} \cmidrule(lr){5-5} \cmidrule(lr){6-6} \cmidrule(lr){7-7} \cmidrule(lr){8-8} \cmidrule(lr){9-9} \cmidrule(lr){10-10}
Rectified Diffusion & 30.14 & 18.70 & 12.08 & 7.68 & 29.00 & 37.66 & 13.85 & 33.77 & 48.42 \\
RF w/o Reweighting & 25.38 & 16.11 & 10.88 & 7.25 & 27.5 & 38.24 & 14.63 & 34.34 & 47.98 \\
RF w/ LogNorm & 25.75 & 16.37 & 11.08 & 7.39 & 27.73 & 38.42 & 14.76 & 34.54 & 48.15 \\
RF w/ ModeSample & 25.11 & 15.92 & 10.74 & 7.16 & 27.38 & 38.27 & 14.61 & 34.41 & 48.59 \\
RF w/ CosMap & 25.20 & 15.97 & 10.77 & 7.18 & 27.32 & 38.23 & 14.57 & 34.37 & 47.77 \\
CurveFlow & \textbf{30.32} & \textbf{19.33} & \textbf{12.99} & \textbf{8.64} & \textbf{30.41} & \textbf{39.92} & \textbf{15.73} & \textbf{35.96} & \textbf{50.18} \\
\bottomrule
\end{tabular}}
\end{table*}

\begin{table}[t]
\centering
% \captionsetup{width=.8\linewidth}
\scriptsize
\caption{Performance of generated images produced by various loss reweighting methods on the validation set (5K samples) of COCO17.
}
% \vspace{-2ex}
\label{tab:perf_gen_coco17}
\adjustbox{width=.4\textwidth}{
\begin{tabular}{lccc}
\toprule
\textbf{Method} & \textbf{FID} {$\downarrow$} & \textbf{IS} {$\uparrow$} & \textbf{CLIPScore} {$\uparrow$}  \\
\cmidrule(lr){1-1} \cmidrule(lr){2-2} \cmidrule(lr){3-3} \cmidrule(lr){4-4} 
Rectified Diffusion & 20.95 & 32.13 & 30.82 \\
RF w/o Reweighting & 21.52 & 33.73 & 32.22 \\
RF w/ LogNorm & 21.55 & 33.80 & \textbf{32.28}  \\
RF w/ ModeSample & 21.98 & 33.20 & 32.19  \\
RF w/ CosMap &  21.70 & \textbf{33.95} & 32.24  \\
CurveFlow & \textbf{20.57} & 33.26 & 31.90 \\
\bottomrule
\end{tabular}}
\end{table}

\begin{table}[t]
\centering
% \captionsetup{width=.8\linewidth}
\scriptsize
\caption{Performance of generated images produced by various loss reweighting methods on the validation set (30K samples) of COCO14.
}
% \vspace{-2ex}
\label{tab:perf_gen_coco14}
\adjustbox{width=.4\textwidth}{
\begin{tabular}{lccc}
\toprule
\textbf{Method} & \textbf{FID} {$\downarrow$} & \textbf{IS} {$\uparrow$} & \textbf{CLIPScore} {$\uparrow$}  \\
\cmidrule(lr){1-1} \cmidrule(lr){2-2} \cmidrule(lr){3-3} \cmidrule(lr){4-4} 
Rectified Diffusion & 10.47 & 35.71 & 30.80 \\
RF w/o Reweighting & 11.08 & 38.12 & 32.27 \\
RF w/ LogNorm & 11.11 & 38.18 & \textbf{32.28} \\
RF w/ ModeSample & 11.48 & \textbf{38.55} & 32.23 \\
RF w/ CosMap & 11.17 & 38.43 & 32.25 \\
CurveFlow & \textbf{10.44} & 38.29 & 31.92  \\
\bottomrule
\end{tabular}}
\end{table}

\noindent\textbf{Dataset}.
The Common Objects in Context (COCO) dataset~\cite{lin2014microsoft} is one of the most widely used benchmark datasets in computer vision research. It provides a comprehensive collection of images for tasks including object detection, segmentation, and captioning. For our experiments, we utilize a training set containing approximately 11K images. To evaluate our method, we employ two different validation sets: the COCO17 validation set consisting of 5K images and the larger COCO14 validation set with around 30K images.
The two validation sets serve complementary purposes in our evaluation protocol. The 5K COCO17 validation set follows the latest COCO evaluation standards and enables direct comparison with contemporary methods. Meanwhile, the 30K COCO14 validation set offers a more extensive evaluation corpus, allowing us to assess our model's performance across a broader range of scenarios and edge cases. This dual validation approach provides a comprehensive understanding of our method's strengths and limitations.

% We evaluate our method on COCO~\cite{liang2024rich}, a novel dataset containing rich human feedback annotations on 18K generated images from the Pick-a-Pic dataset. The dataset consists of 16K training samples, 1K validation samples, and 1K test samples. Each image-text pair is annotated by three human annotators with: (1) point annotations highlighting regions of implausibility/artifacts and text-image misalignment, (2) labeled misaligned keywords in the text prompts, and (3) four fine-grained scores (ranging from 1-5) evaluating plausibility, text-image alignment, aesthetics, and overall quality. The annotations are consolidated through averaging scores, majority voting for misaligned keywords, and averaging heatmaps generated from the point annotations. The dataset is balanced across different content categories including humans, animals, objects, indoor scenes, and outdoor scenes, with approximately 60\% being photorealistic images.

\noindent\textbf{Models \& Metrics}.
Following \cite{esser2024scaling}, we include the three sampling methods, \ie, LogNorm, ModeSample, and CosMap for comparison purposes. CurveFlow is implemented based on the framework of \cite{esser2024scaling}. In addition, we compare against rectified diffusion (Phased, the best model in \cite{rectified_diffusion_2025}), which employs a non-straight trajectory.
To gain a comprehensive understanding of CurveFlow, we conduct extensive evaluations on image-to-text semantic consistency (BLEU~\cite{papineni2002bleu}, METEOR~\cite{banerjee2005meteor}, ROUGE~\cite{lin2004rouge}, and CLAIR~\cite{chan2023clair}), image quality (FID~\cite{heusel2017gans} and IS~\cite{salimans2016improved}), and text-image alignment (CLIPScore~\cite{hessel2021clipscore}). we employ BLIP v2 \cite{li2023blip} to generate captions based on the synthesized images, then compute the NLP metrics between these generated captions and the reference captions. By benchmarking on these standardized COCO splits and utilizing multiple complementary metrics, we ensure fair comparison with state-of-the-art approaches while providing a holistic assessment of our method's performance.

\begin{figure*}[!t]
\centering
\includegraphics[width=.9\linewidth]{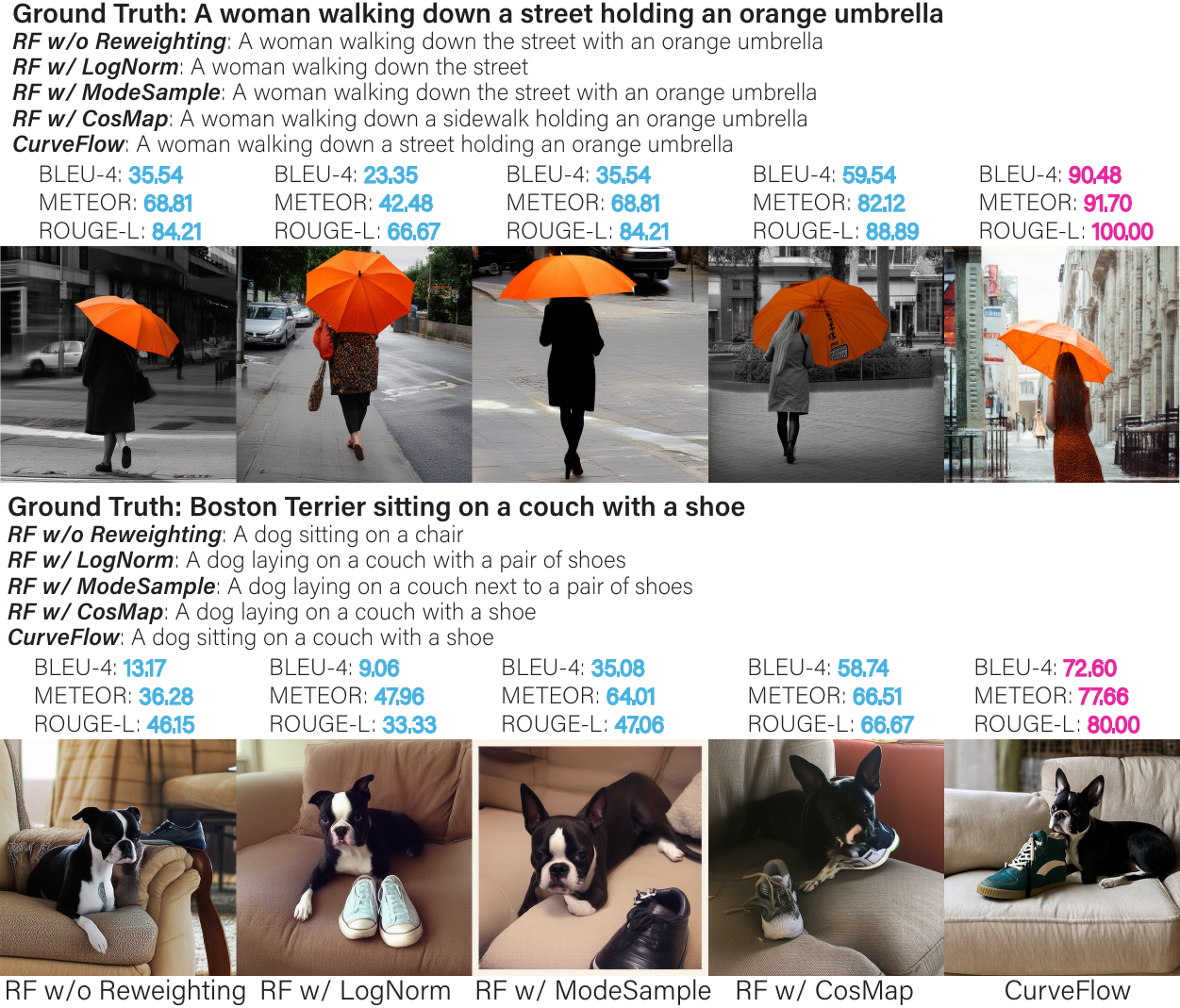}
\caption{Qualitative comparison showing that CurveFlow generates captions with superior semantic accuracy and detail alignment compared to RF variants.}
\label{fig:qual_comparison}
\end{figure*}

\noindent\textbf{Performance}.
The experimental results show a consistent ranking among the various loss reweighting strategies within the Rectified Flow framework \cite{esser2024scaling}, with minor but measurable differences in performance. Among the RF variants, RF w/ LogNorm achieves the best overall performance on both COCO17 and COCO14, slightly outperforming RF w/o Reweighting, RF w/ CosMap, and RF w/ ModeSample on most caption-based metrics. For instance, on COCO17 (Table~\ref{tab:perf_caption_coco17}), RF w/ LogNorm obtains a BLEU-4 of 7.49 and CLAIR score of 62.77, surpassing RF w/o Reweighting (7.35, 62.69), RF w/ CosMap (7.45, 62.56), and RF w/ ModeSample (7.36, 62.31). A similar trend holds on COCO14 (Table~\ref{tab:perf_caption_coco14}), where RF w/ LogNorm leads in BLEU-4 (7.39), METEOR (27.73), and CLAIR (48.15). This suggests that logarithmic normalization of loss weights provides a more stable and effective gradient signal during training compared to mode-based sampling or cosine similarity weighting, which may introduce bias or insufficient dynamic range. Despite these refinements, all RF variants underperform significantly compared to Rectified Diffusion and, more notably, CurveFlow, particularly in higher-order n-gram scores (BLEU-3, BLEU-4) and CLAIR, indicating limitations in capturing complex semantic structures due to their shared reliance on linear trajectories.

In contrast, Rectified Diffusion, which relaxes the strict linearity of RF by learning more flexible, non-linear trajectories, demonstrates a clear advantage over all RF variants, serving as a strong baseline that validates the importance of non-linear flow paths. On COCO17, Rectified Diffusion achieves 29.44 BLEU-1 and 8.15 BLEU-4, outperforming all RF methods by a notable margin (e.g., +3.84 in BLEU-1 over RF w/o Reweighting), and attains a CLAIR score of 61.08, though still below the best RF variant (62.77). This indicates that while Rectified Diffusion improves semantic consistency, it may trade off some instruction following for broader caption similarity. However, CurveFlow surpasses both Rectified Diffusion and all RF variants across nearly all metrics, achieving 29.54 BLEU-1 and 64.26 CLAIR on COCO17—demonstrating that curvature-aware trajectory modeling enhances both fluency and precise semantic alignment. Importantly, this gain is achieved while maintaining superior image quality, with CurveFlow attaining the best FID on both datasets (20.57 on COCO17, 10.44 on COCO14), surpassing even Rectified Diffusion (20.95 and 10.47). These results confirm that CurveFlow's curvature regularization enables smoother, geometrically informed transitions, yielding images that are not only visually faithful but also semantically aligned with input instructions.

\begin{figure}[htbp]
\captionsetup[subfigure]{labelformat=empty}
  \centering
  \subfloat[]{
    \includegraphics[width=0.21\textwidth]{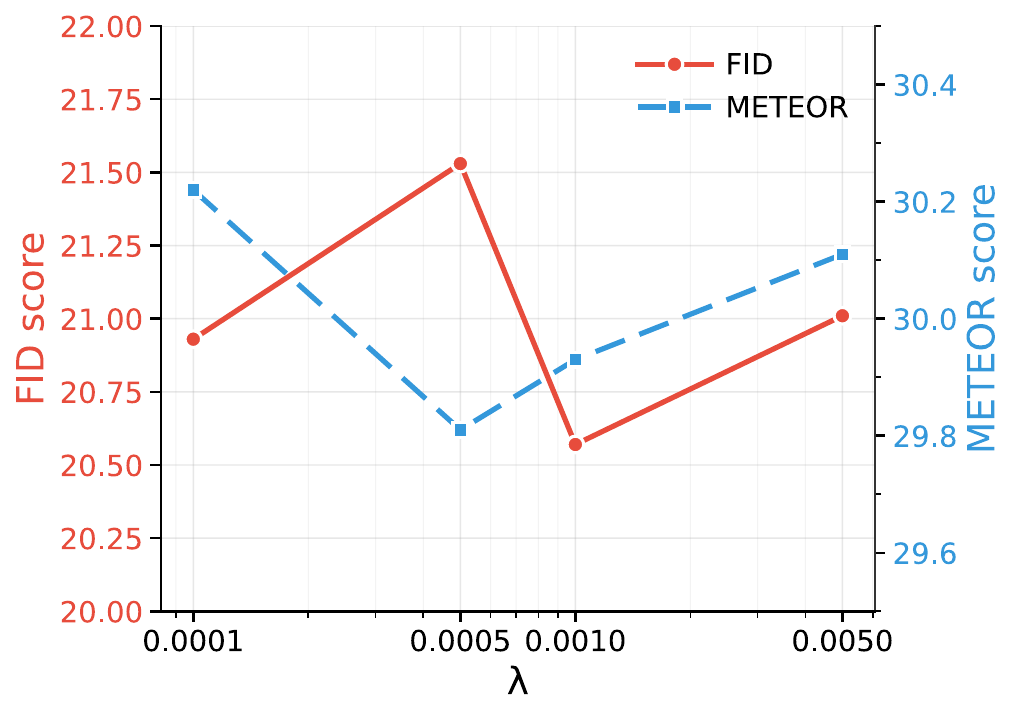}
  }
  \hfill
  \subfloat[]{
    \includegraphics[width=0.21\textwidth]{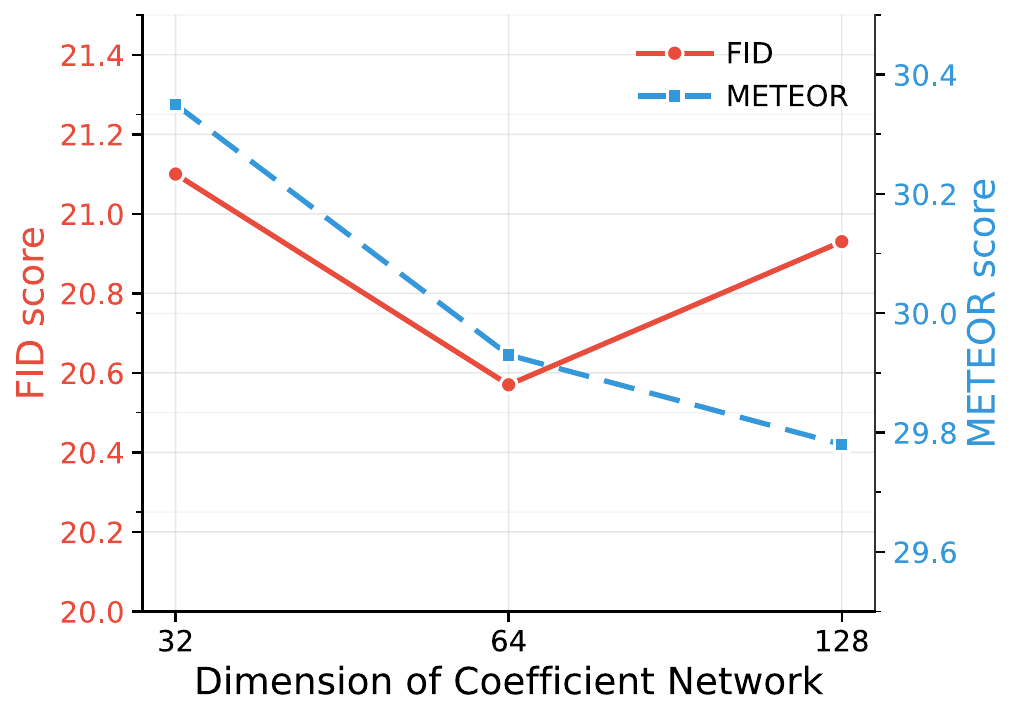}
  }
 \vspace{-4ex}
  \caption{Impact of $\lambda$ (left) and the dimensions of coefficient networks (right) on FID and METEOR.}
  \label{fig:ablation}
\end{figure}

\noindent\textbf{Ablation Study on $\lambda$}
Our experimental results, as visualized in Figure~\ref{fig:ablation}, demonstrate the non-trivial impact of the regularization parameter $\lambda$ on model performance. When examining both the METEOR and ROUGE-1 metrics across varying regularization strengths ($\lambda \in \{0, 0.001, 0.01, 0.1, 1\}$), we observe similar performance trajectories for both evaluation criteria. At $\lambda = 0$ (no regularization), the model achieves baseline performances of 29.08 (METEOR) and 37.95 (ROUGE-1), indicating potential underfitting. As $\lambda$ increases incrementally, we note a monotonic improvement in both metrics until reaching optimal values at $\lambda = 0.1$, with METEOR and ROUGE-1 scores of 30.52 and 40.17, respectively—representing relative improvements of 4.95\% and 5.86\% over the baseline. This indicates an optimal trade-off between underfitting and overfitting at this regularization value. The logarithmic x-axis highlights the sensitivity of model performance to small changes in $\lambda$, particularly in the range of 0.01-0.1. These results demonstrate the importance of proper regularization tuning for maximizing the semantics consistency in the generated images as measured by these complementary evaluation metrics.

% \begin{table*}[t]
% \centering
% % \captionsetup{width=.8\linewidth}
% \scriptsize
% \caption{Performance of generated images produced by various loss re-weighting methods on RHF18K cite XXX on human feedback prediction. Stable Diffusion v3.5 is used for the experiment. 
% }
% % \vspace{-2ex}
% \label{tab:perf_hf_prediction}
% \adjustbox{width=.8\textwidth}{
% \begin{tabular}{lcccc}
% \toprule
% \textbf{Method} & \textbf{Avg Plausibility} {$\uparrow$} & \textbf{Avg Alignment} {$\uparrow$} & \textbf{Avg Aesthetics} {$\uparrow$} & \textbf{Avg Overall} {$\uparrow$} \\
% \cmidrule(lr){1-1} \cmidrule(lr){2-2} \cmidrule(lr){3-3} \cmidrule(lr){4-4} \cmidrule(lr){5-5}
% No Fine-Tuning & 0.7558 & 0.6793 & 0.7558 & 0.6588 \\
% Fine-Tuning w/o Weighting & 0.7372 & 0.6713 & 0.7415 & 0.6447 \\
% Fine-Tuning w/ LogNorm Sampling & 0.7433 & 0.6769 & 0.7471 & 0.6512 \\
% Fine-Tuning w/ Mode Sampling & 0.7450 & 0.6745 & 0.7477 & 0.6505 \\
% Fine-Tuning w/ CosMap & 0.7409 & 0.6740 & 0.7445 & 0.6479 \\
% Fine-Tuning w/ Proposed & 0.7708 & 0.6837 & 0.7682 & 0.6694\\
% \bottomrule
% \end{tabular}}
% \end{table*}

\noindent\textbf{Qualitative Analysis}. In Figure~\ref{fig:qual_comparison} provides a qualitative illustration of how CurveFlow improves semantic accuracy and instruction following compared to various loss reweighting variants of RF and Rectified Diffusion. In the first example, the ground truth describes a woman walking down a street holding an orange umbrella. While several methods (e.g., RF w/ LogNorm, RF w/ ModeSample) generate plausible captions, only CurveFlow exactly matches the ground truth, capturing both the action (``walking") and the specific object (``orange umbrella"). Notably, RF w/ LogNorm completely omits the umbrella, a key visual detail, resulting in a significant drop in BLEU-4 (23.35 vs. 90.48) and ROUGE-L (66.67 vs. 100.00). This highlights the tendency of baseline RF models to generate generic or incomplete descriptions despite reasonable visual fidelity.

In the second example, the input image depicts a Boston Terrier sitting on a couch with a shoe. CurveFlow again produces the most accurate description, correctly identifying the dog breed (``Boston Terrier"), posture (``sitting"), and object (``a shoe"). In contrast, all RF variants and Rectified Diffusion fail to capture at least one critical detail: most describe the dog as ``laying" instead of ``sitting", and many incorrectly refer to ``a pair of shoes" or omit the object entirely. These hallucinations and inaccuracies are reflected in substantially lower metric scores—for instance, RF w/ CosMap achieves only 66.67 ROUGE-L and 58.74 BLEU-4, compared to CurveFlow's 88.00 and 72.60, respectively.
This consistent advantage demonstrates that curvature-aware trajectory modeling in CurveFlow enables finer-grained semantic control and better instruction following. By learning smoother, geometrically informed paths through the data manifold, CurveFlow avoids abrupt or unnatural transitions—effectively staying within high-probability regions of the data space—thereby preserving critical visual details that are often lost under the rigid linear flows of RF and its variants.
% The consistent superiority of CurveFlow across both examples demonstrates that curvature-aware trajectory modeling enables more precise alignment between visual content and textual semantics. CurveFlow stays on a more natural, high-probability path. The model preserves semantic structure and fine-grained details that might otherwise be lost or distorted under a linear flow.
% By learning smoother, geometrically informed paths in the data manifold, CurveFlow avoids low-probability transitions that may lead to semantic drift, thereby enhancing fine-grained detail preservation and overall instructional compliance.

\noindent\textbf{Ablation Study}.
As shown in Figure~\ref{fig:ablation}, the choice of $\lambda = 0.001$ and coefficient network dimension 64 represents an optimal trade-off between image quality, semantic alignment, and training stability. As demonstrated in the left panel, setting $\lambda = 0.001$ achieves the lowest FID score (indicating best sample quality) and relatively high METEOR score (indicating strong semantic consistency), while larger values ($\lambda \geq 0.01$) degrade performance due to over-regularization, which overly constrains the trajectory flexibility and hinders expressive modeling. Conversely, $\lambda = 0$ (no curvature regularization) results in higher FID and lower METEOR, confirming that some curvature control is essential for high-fidelity and instruction-compliant generation. The right panel further shows that increasing the network dimension from 32 to 64 significantly improves both FID and METEOR, indicating that a higher-dimensional representation is necessary to model the complex dynamics of $a_\phi(t)$ and $b_\psi(t)$ effectively. However, further increasing the dimension to 128 leads to a performance drop, likely due to overfitting and optimization instability. Therefore, dimension 64 provides sufficient model capacity without compromising generalization, making it the optimal choice. 

\noindent\textbf{Computational Time}.
Both CurveFlow and the various RF variants achieve comparable inference times, as all models are based on the same ODE-solving framework and require an identical number of sampling steps. In terms of training efficiency, CurveFlow demonstrates a favorable trade-off between computational cost and performance. Specifically, on 4 NVIDIA H100 GPUs, RF w/o Reweighting requires 84 GPU hours (4×21), serving as the baseline. The reweighting strategies introduce additional overhead: RF w/ ModeSample takes 91.6 GPU hours (4×22.9), and RF w/ CosMap incurs the highest cost at 106.4 GPU hours (4×26.6), due to the computational complexity of cosine similarity calculations during loss weighting. In contrast, CurveFlow, which incorporates curvature regularization and learns dynamic scaling functions $a_\phi(t)$ and $b_\psi(t)$, requires 92.4 GPU hours (4×23.1)—only slightly more than RF w/ ModeSample and significantly less than RF w/ CosMap. 

%% file: sec/5_conclusion.tex
\section{Conclusion}
We propose CurveFlow, a curvature-guided flow matching framework that overcomes the geometric limitations of rectified flow by learning smooth, non-linear trajectories between data and noise. Experiments on MS COCO 2014 and 2017 show that CurveFlow achieves state-of-the-art performance in text-to-image generation, significantly improving instructional compliance—evidenced by superior scores in BLEU, METEOR, ROUGE, and CLAIR—while maintaining competitive image quality with lower FID scores. These results demonstrate that modeling curvature-aware paths is essential for generating images that faithfully align with complex textual instructions.